\documentclass[10pt,twocolumn,letterpaper]{article}

\usepackage{iccv}
\usepackage{times}
\usepackage{epsfig}
\usepackage{graphicx}
\usepackage{amsmath}
\usepackage{amssymb}
\usepackage[ruled,boxed,noline]{algorithm2e}
\usepackage{multirow}
\usepackage{subcaption}
\usepackage{bbm}
\usepackage[font={small}]{caption}
\allowdisplaybreaks[4]
\usepackage[pagebackref=true,breaklinks=true,letterpaper=true,colorlinks,bookmarks=false]{hyperref}

\iccvfinalcopy % *** Uncomment this line for the final submission

\def\iccvPaperID{****} % *** Enter the ICCV Paper ID here

\ificcvfinal\fi
\begin{document}

%%%%%%%%% TITLE
%\title{Unconstrained Video-Based Face Recognition by Uncertainty-Gated Graph}
\title{Uncertainty Modeling of Contextual-Connections between Tracklets for Unconstrained Video-based Face Recognition}

\author{Jingxiao Zheng\textsuperscript{1}
\and
Ruichi Yu\textsuperscript{1}\thanks{Currently working in Waymo.}\\
\and
Jun-Cheng Chen\textsuperscript{2}\\
\and
Boyu Lu\textsuperscript{1}\\
\and
Carlos D. Castillo\textsuperscript{1}\\
\and
Rama Chellappa\textsuperscript{1}\\
\textsuperscript{1} UMIACS, University of Maryland, College Park\quad\textsuperscript{2} CITI, Academia Sinica, Taiwan\\
{\tt\small \{jxzheng, yrcbsg\}@umiacs.umd.edu, pullpull@citi.sinica.edu.tw, \{bylu, carlos, rama\}@umiacs.umd.edu}
}
\maketitle
\ificcvfinal\thispagestyle{empty}\fi

%%%%%%%%% ABSTRACT
\begin{abstract}
%Unconstrained video-based face recognition is a challenging problem due to the significant within-video variations like pose, occlusion and blur. An effective idea to improve the performance is using variation-robust context information, such as body appearance, to connect the samples and propagate the identity information from high-quality faces to low-quality ones. However, previous methods usually use fixed connections during information propagation, which may lead to errors if the connections are not correct. In this paper, we present a method called Uncertainty-Gated Graph to model the uncertainty of the connections built by these context information. Specifically, we add additional gates on connection graph edges, which enable the edge weights to be adaptively adjusted according to the strength of identity correlations between samples. In addition, the gates allow two types of information to propagate between samples simultaneously. The proposed model can be directly used without training, or plugged onto other network architecture for supervised training. The proposed method also achieves state-of-art results in the recently released challenging Cast Search in Movies and IARPA Janus Surveillance Video Benchmark datasets. 

Unconstrained video-based face recognition is a challenging problem due to significant within-video variations caused by pose, occlusion and blur. 
To tackle this problem, an effective idea is to propagate the identity from high-quality faces to low-quality ones through contextual connections, which are constructed based on context such as body appearance.
However, previous methods have often propagated erroneous information due to lack of uncertainty modeling of the noisy contextual connections.
In this paper, we propose the Uncertainty-Gated Graph (UGG), which conducts graph-based identity propagation between tracklets, which are represented by nodes in a graph. UGG explicitly models the uncertainty of the contextual connections by adaptively updating the weights of the edge gates according to the identity distributions of the nodes during inference. 
UGG is a generic graphical model that can be applied at only inference time or with end-to-end training.
We demonstrate the effectiveness of UGG with state-of-the-art results in the recently released challenging Cast Search in Movies and IARPA Janus Surveillance Video Benchmark dataset. 
\end{abstract}
\vspace{-15pt}
%%%%%%%%% BODY TEXT
\section{Introduction}\label{sec:intro}
Unconstrained video-based face recognition has been an active research topic for decades in computer vision and biometrics. In a wide range of its applications, such as visual surveillance, video content analysis and access control, the task is to match the subjects in unconstrained probe videos to pre-enrolled gallery subjects, which are represented by still face images. Although recent advances of deep convolutional neural network (DCNN)-based methods have achieved comparable or superior performance to human in still-image based face recognition \cite{Taigman2014, Parkhi15, Schroff2015, JC15, crystal, Sun2014, Sun2014Sparse,Ding2016, arcface}, unconstrained video-based face recognition still remains a challenging problem due to significant facial appearance variations caused by pose, motion blur, and occlusion.

\begin{figure}[t]
\centering
\includegraphics[width=\linewidth]{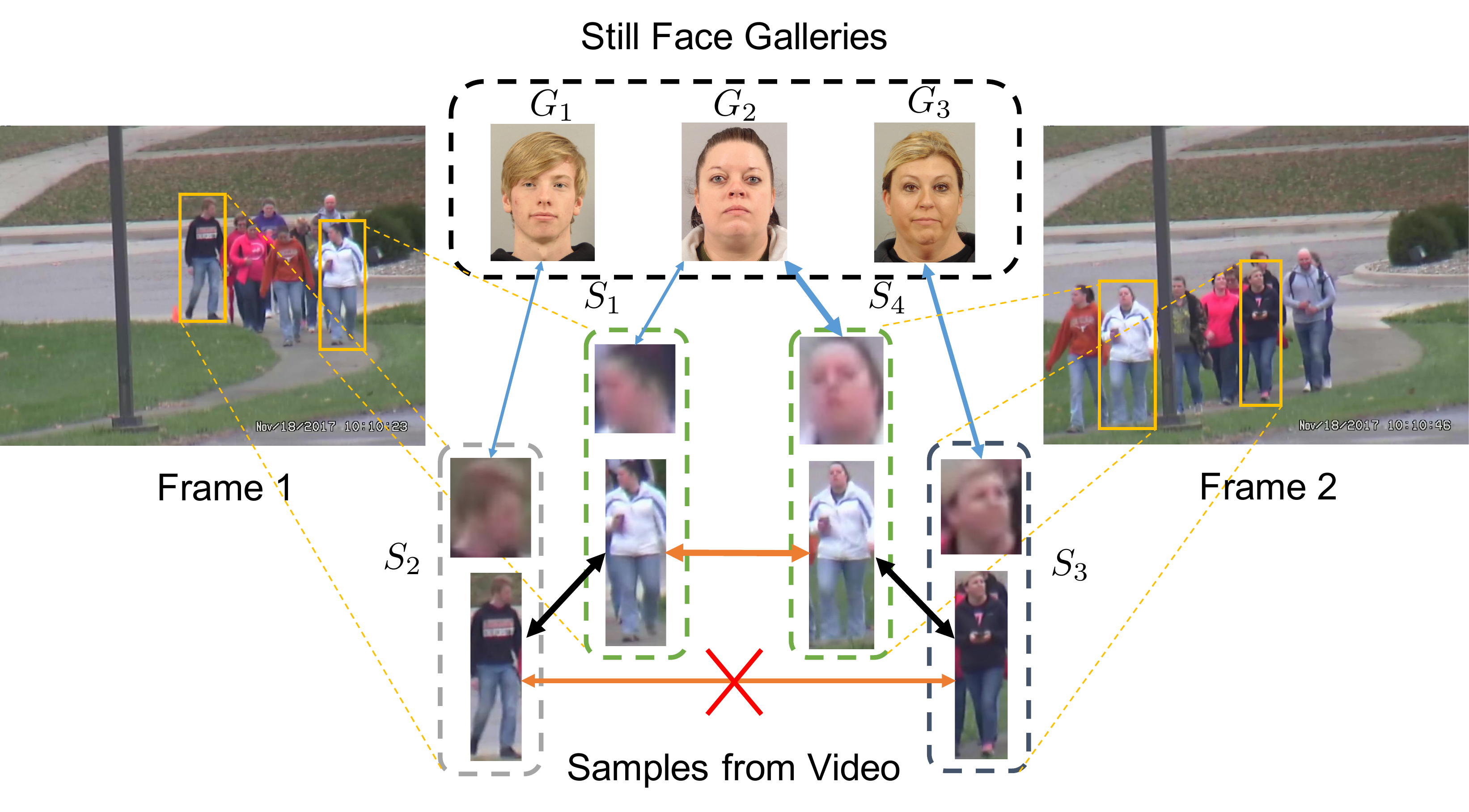}
\caption{An example of video-based face recognition problem consisting of three still face gallery subjects and four samples from the videos. {\color[rgb]{1,0.5,0}Orange arrows} show positive connections from body appearance similarity. \textbf{Black arrows} indicate negative connections constructed from co-occurrence information. {\color{blue}Blue arrows} represent the facial similarities to the ground truth galleries. The thicker the arrows, the stronger the connections. The {\color{red}red cross} indicates an misleading connection. A graph with fixed connections may propagate erroneous information through these misleading connections. (The figure is best viewed in color.)}
\label{fig:relationship}
\vspace{-15pt}
\end{figure}
To fill the performance gap between face recognition in still-images and unconstrained videos, one possible solution is to train a video-specific model with large amount of training data, which is difficult and costly to collect. Another effective idea is to leverage the well-studied image-based face recognition methods to first identify video faces with limited variations, then utilize some video contextual information, such as body appearance and spatial-temporal correlation between person instances, to propagate the identity information from high-quality faces to low-quality ones. For instance, in Figure~\ref{fig:relationship}, by utilizing the body appearance, we may propagate the identity information obtained from frontal face $S_4$ to the profile face $S_1$, which is very difficult to recognize individually.

The above idea has been explored using graph-based approaches \cite{personsearch, crf_fa, reid_graph}. Graphs are constructed with nodes to represent one or more frames (tracklets) of person instances and edges to connect tracklets. However, a major limitation of these approaches is that their graphs are pre-defined and the edges are fixed during information propagation. A misleading connection may propagate erroneous information. As shown in Figure~\ref{fig:relationship}, these methods may propagate the identity information between $S_2$ and $S_3$ based on their similar body appearance, which might lead to erroneous propagation.

To address the problem, we propose a graphical-model-based framework called Uncertainty-Gated Graph (UGG) to model the uncertainty of connections built using contextual information. We formulate UGG as a conditional random field on the graph with additional gate nodes introduced on the connected graph edges. With a carefully designed energy function, the identity distribution of tracklets\footnote{We follow the same definition of tracklets with \cite{personsearch}.} is updated by the information propagated through these gate nodes during inference. In turn, these gate nodes are adaptively updated according to the identity distributions of the connected tracklets. The uncertainty gate nodes consist of two types of gates: positive gates that control the confidence of the positive connections (encourage the connected pairs to have the same identity) and negative gates that control negative ones (discourage pairs to have the same identity). It is worth noting that negative connections can significantly contribute to performance improvements by discouraging similar identity distribution between clearly distinct subjects, e.g., two people in the same frame\footnote{In Figure~\ref{fig:relationship}, the co-occurrence of $S_3$ and $S_4$ in the same frame of the video is a strong prior to indicate their different identities.}. Explicitly modeling positive/negative information separately allows our model to consider different contextual information in challenging conditions, and leads to improved uncertainty modeling. %Considering the uncertainty of the connection between nodes, random fields are used to model the opening status of the gates that control the connections. Energy function is designed such that edge weights can be appropriately updated according to the identity distributions of neighboring nodes during inference.

Our approach can be directly applied at inference time, or plugged onto an end-to-end network architecture for supervised and semi-supervised training. The proposed method is evaluated on two challenging datasets, the Cast Search in Movies (CSM) dataset \cite{personsearch} and the IARPA Janus Surveillance Video Benchmark (IJB-S) dataset \cite{ijbs} with superior performance compared to existing methods.

The main contributions of this paper are summarized as follows:
\vspace{-5pt}
\begin{itemize}
\item We propose the Uncertainty-Gated Graph model for video-based face recognition by explicitly modeling the uncertainty
of connections between tracklets using uncertainty gates over graph edges. The tracklets and gates are updated jointly and possible connection errors might be corrected during inference.
\item We utilize both positive and negative connections for information propagation. Despite its effectiveness, negative connections were often ignored in previous approaches for unconstrained face recognition.
\item The proposed method is efficient and flexible. It can either be used at inference time without supervision, or be considered as a trainable module for supervised and semi-supervised training.

%\item We achieve state-of-the-art results in two challenging datasets, the CSM dataset and the IJB-S dataset.
\vspace{-10pt}
\end{itemize}

%The rest of this paper is organized as follows: in Section~\ref{sec:related} we briefly review some related works. In Section~\ref{sec:method} we introduce the proposed method in detail. In Section~\ref{sec:exp}, we discuss the implementation details and present the experimental results on two datasets. Finally, conclusions are presented in Section~\ref{sec:conclusion}.

\begin{figure*}[t]
\centering 
\includegraphics[width=0.9\linewidth]{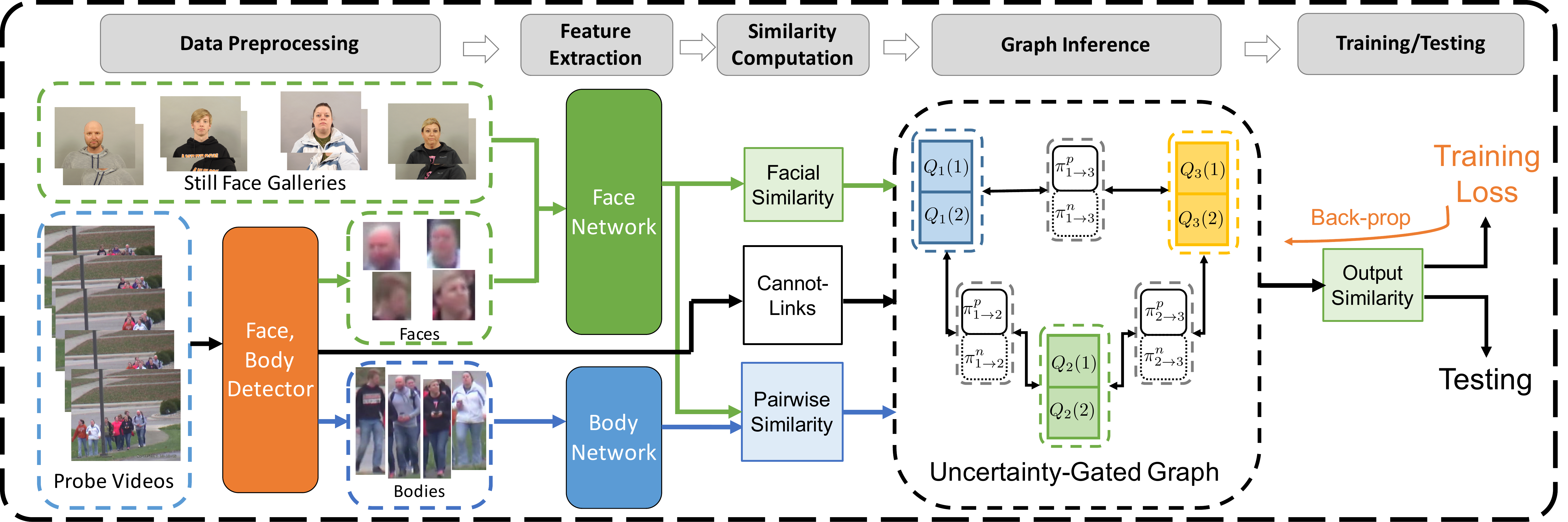}
\vspace{-5pt}
\caption{Overview of the proposed method. Given still face galleries and probe videos, we first detect all the faces and corresponding bodies from the videos. Faces are associated into tracklets by a tracker. Face features for galleries and tracklets, and body features for tracklets are extracted by corresponding networks. Similarities are computed from these flattened features. Facial and body similarities, together with cannot-link constrains from the detection information are fed into the proposed UGG model. After inference, the output is used for testing, or generating the loss for end-to-end training.}
\label{fig:diagram}
\vspace{-15pt}
\end{figure*}
\section{Related Works}\label{sec:related}
%In this section, we briefly review some relevant works:
%
\textbf{Deep Learning for Face Recognition:}
Deep learning is widely used for face recognition tasks as it has demonstrated significant performance improvements. %Taigman \emph{et al.} \cite{Taigman2014} learned a DCNN model on the frontalized faces generated from 3D shape models. 
Sun \emph{et al.} \cite{Sun2014, Sun2014Sparse} achieved results surpassing human performance on the LFW dataset \cite{LFW}. %Schroff \emph{et al.} \cite{Schroff2015} adopted the GoogLeNet trained for object recognition to face recognition and trained on a large-scale unaligned face dataset.
Parkhi \emph{et al.} \cite{Parkhi15} achieved impressive results for face verification. Chen \emph{et al.} \cite{JC15, jc_ijcv} reported very good performance on IJB-A, JANUS CS2, LFW and YouTubeFaces \cite{YTF} datasets. Ranjan \emph{et al.} \cite{crystal} achieved good performance on IJB-C\cite{ijbc}. Zheng \emph{et al.} \cite{video_journal} achieved good performance on video face datasets including IJB-B \cite{ijbb} and IJB-S \cite{ijbs}. \cite{arcface} presents a recent face recognizer with state-of-the-art performance.

\textbf{Label Propagation:}
Label propagation \cite{LP1} has many applications in computer vision. Huang \emph{et al.} \cite{personsearch} proposed an approach for searching person in videos using a label propagation scheme instead of trivial label diffusion. Kumar \emph{et al.} \cite{LP_ICPR} proposed a video-based face recognition method by selecting key-frames and propagating the labels on key-frames to other frames. Sheikh \emph{et al.} \cite{LP_ECCVW} used label propagation to reduce the runtime for semantic segmentation using random forests. Tripathi \emph{et al.} \cite{LP_WACV} introduced a label propagation-based object detection method.

\textbf{Conditional Random Field:}
The Conditional Random Field (CRF) \cite{CRF} is a commonly used probabilistic graphical models in computer vision research. Kr\"{a}henb\"{u}hl  \emph{et al.} \cite{crf2011} is one of the early researchers to use CRF for semantic segmentation. Chen \emph{et al.} \cite{deeplab, crfsemantic} proposed a DCNN-based system for semantic segmentation and used a CRF for post-processing. Zheng \emph{et al.} \cite{crfasrnn} further introduced an end-to-end framework of a deep network with a CRF module for semantic segmentation. Du \emph{et al.} \cite{crf_fa} used a CRF to solve the face association problem in unconstrained videos.

\textbf{Graph Neural Networks:}
A Graph Neural Network (GNN) \cite{GNN, DeepCNG} is a neural network combined with graphical models such that messages are passed in the graph to update the hidden states of the network. Shen \emph{et al.} \cite{reid_graph} used a GNN for person re-identification problem. %In the GNN, the graph is built on the scores of probe-gallery pairs and the edges are determined by the similarity of gallery-gallery pairs.
Hu \emph{et al.} \cite{label_relation} introduced a structured label prediction method based on a GNN, which allows positive and negative messages to pass between labels guided by external knowledge. But the graph edges are fixed during testing. Wang \emph{et al.} \cite{zero-shot} introduced a zero-shot learning method %by distilling information from word embeddings and knowledge graphs
using stacked GNN modules. Lee \emph{et al.} \cite{ML_zero-shot} proposed another multi-label zero-shot learning method by message passing in a GNN based on knowledge graphs. 

Most of the graph-based methods mentioned above only allow positive messages to pass in the graph, and all of them rely on graphs with fixed edges during testing.% The corresponding improvement of the proposed method is introduced in Section~\ref{sec:method}.
\section{Proposed Method}\label{sec:method}
The overview of the method is shown in Figure~\ref{fig:diagram}. For each probe video, faces are detected and associated into tracklets. Initial facial similarities between gallery images and probe tracklets are computed by a still face recognizer. Connections between tracklets are generated based on the similarity of their facial, body appearances and their spatial-temporal relationships. Then, we build the UGG where these tracklets and connections act as nodes and edges. The connections between tracklets are modeled as uncertainty gates between nodes. The inference can be efficiently implemented by message passing to optimize the energy function of the UGG module.

\subsection{Problem Formulation}\label{subsec:formulation}
For a video-based face recognition problem, suppose we have $C$ gallery subjects and a probe video. The faces in this video are first detected and tracked into $N$ tracklets. For each tracklet, we compute $C$ similarity scores to gallery subjects.% and the best matched gallery subject can be found by ranking the scores.

Suppose we are given the gallery-to-tracklet similarity $\mathbf{S}^{gt}= \begin{bmatrix}s_{li}^{gt}\end{bmatrix}\in \mathbb{R}^{C\times N}$ and the tracklet-to-tracklet similarity $\mathbf{S}^{tt}=\begin{bmatrix}s_{ij}^{tt}\end{bmatrix}\in\mathbb{R}^{N\times N}$, where $s_{li}^{gt}$ is the similarity between the gallery $l$ and the tracklet $i$, $s_{ij}^{tt}$ is the similarity between tracklet $i$ and $j$. Furthermore, a cannot-link matrix $\mathbf{L}^{tt}=\begin{bmatrix}L_{ij}^{tt}\end{bmatrix}\in\{0, 1\}^{N\times N}$ is given such that
\begin{equation}
\small
L_{ij}^{tt} = \begin{cases}
1 &\text{identities of tracklet $i$ and $j$ are different}\\
0 & \text{no constraint}
\end{cases}
\end{equation}
% A cannot-link exists between tracklet $i$ and $j$ if they absolutely do not belong to same gallery subject.

Here, $\mathbf{S}^{gt}$ provides prior identity information, $\mathbf{S}^{tt}$ provides the positive contextual information between tracklets and $\mathbf{L}^{tt}$ provides the negative contextual information. %For example, two distinct faces appear in the same video frame, which always belong to different identities.
By combining these information, the output gallery-to-tracklet similarity is computed as
\begin{equation}
\small
\tilde{\mathbf{S}}^{gt}=UGG(\mathbf{S}^{gt}, \mathbf{S}^{tt}, \mathbf{L}^{tt})\in\mathbb{R}^{C\times N}
\end{equation}
where $UGG(\cdot)$ is a function based on the proposed Uncertainty-Gated Graph. In the following sections, we introduce the model in detail.

\subsection{Uncertainty-Gated Graph}
%The proposed graph model is illustrated in Figure~\ref{fig:graph}.
First, given a video with $N$ tracklets detected, a graph $\mathcal{G}=(\mathcal{V}, \mathcal{E})$ is built where each node corresponds to a tracklet. Node $i$ is only connected to its neighbors $\mathcal{N}(i)$. Based on the graph $\mathcal{G}$, we define a random field $\mathbf{X}=\{X_1,\dots,X_N\}$ associated to nodes $\mathcal{V}$. $X_i\in\mathcal{L} = \{1, \dots, C\}$ is the label variable of tracklet $i$. $X_i=l$ means gallery subject $l$ is assigned to tracklet $i$. We call these nodes as \textit{sample nodes}.

We further add \textit{gates nodes} to each of the edges in $\mathcal{E}$ attached with a random field $\mathbf{Y} = \{Y_{i\to j}^p, Y_{i\to j}^n\}$. In each gate node $i\to j$, we place two gate variables, the \textit{positive gate} $Y_{i\to j}^p\in\{0, 1\}$ and the \textit{negative gate} $Y_{i\to j}^n\in\{0, 1\}$, to control the connections between tracklets $i$ and $j$. %The \textbf{positive gate} $Y_{i\to j}^p=1$ implies that tracklet $i$ and $j$ have the same label and $Y_{i\to j}^p=0$ means weak positive information. Similarly, the \textbf{negative gate} $Y_{i\to j}^n=1$ suggests that tracklet $i$ and $j$ have different labels and $Y_{i\to j}^n=0$ means weak negative information. %Notice that $Y_{i\to j}^p=0$ does not imply a negative relationship, and vice versa.
%We call the proposed model as Uncertainty-Gated Graph (UGG), since it explicitly models the uncertainty of connections between sample nodes as gate variables and controls the information propagation in the graph.

\subsubsection{Energy Function}
% conditioning on the observation $\mathbf{V}$ (the input video in our scenario), we write the posterior probability as a Gibbs distribution
% \begin{equation}
% P(\mathbf{X}, \mathbf{Y}|\mathbf{V})=\frac{1}{Z(\mathbf{V})}\exp(-E(\mathbf{X}, \mathbf{Y}|\mathbf{V}))
% \end{equation}
% where $E(\mathbf{X}, \mathbf{Y}|\mathbf{V})$ is the energy function and $Z(\mathbf{V})$ is the normalization factor. We omit the conditioning on $\mathbf{V}$ in the notation from now on for convenience. The energy function is defined as
The energy function of the UGG module is defined as
\begin{align}
E(\mathbf{x}, \mathbf{y})&=\sum_{i\in\mathcal{V}}\psi_u^x(x_i) + \sum_{i\in\mathcal{V}, j\in\mathcal{N}(i)}[\psi_u^p(y_{i\to j}^p) + \psi_u^n(y_{i\to j}^n)\nonumber\\
&+\psi_t^p(x_i, x_j, y_{i\to j}^p) + \psi_t^n(x_i, x_j, y_{i\to j}^n)]
\end{align}
% By minimizing the energy function $E(\mathbf{x}, \mathbf{y})$, 
% %(i.e. maximizing the posterior $P(\mathbf{x})$ )
% the unary potentials control the label assignment of tracklet nodes and relational nodes independently, while triplet potentials control the interaction between tracklet nodes and relational nodes.

% As we discussed in Section~\ref{sec:intro}, the tracklet-to-tracklet similarity and cannot-link constraints are cues that imply relationships between tracklets. But in principle they are not always reliable. Thus instead of using these relationships as fixed edges between tracklet nodes, we introduce relational nodes and the corresponding potential functions to control the interaction between tracklet nodes. The relational nodes are also adaptively refined by the identity information from the tracklet nodes in return.
The unary potential for tracklet $i$ is defined based on the identity information $\mathbf{S}^{gt}$ as
\begin{equation}
\small
\psi_u^x(x_i=l) = - T_{gt} \cdot s_{li}^{gt}
\end{equation}
where $T_{gt}$ is the temperature factor. The penalty will be low if identity information $s_{li}^{gt}$ is strong. 

We also define the unary potential for the positive gate based on relationship information $\mathbf{S}^{tt}$ as
\begin{equation}
\small
\psi_u^p(y_{i\to j}^p=1) = - T_{tt} \cdot s_{ij}^{tt}
\end{equation}
where $T_{tt}$ is the corresponding temperature factor. Penalty of an open positive gate at edge $i\to j$ will be low if positive connection $s_{ij}^{tt}$ is strong.

The unary potential for the negative gate is defined as
\begin{equation}\label{eq:negative_relation}
\small
\psi_u^n(y_{i\to j}^n=k) = \begin{cases}
0 & \text{if } L_{ij}^{tt}=k\\
+\infty & \text{otherwise}
\end{cases}
\end{equation}
for $k\in\{0, 1\}$. Therefore, opening of the negative gate at node $i\to j$ is determined by the negative connection $L_{ij}^{tt}$.%if a cannot-link exists between pair $(i, j)$, the penalty of choosing $y_{i\to j}^n=0$ will be infinity. Similarly, if there is no cannot-link between pair $(i, j)$, we will never let $y_{i\to j}^n=1$.

The positive triplet potential is defined as
\begin{equation}
\small
\psi_t^p(x_i, x_j, y_{i\to j}^p) =
\begin{cases}
\alpha_p & \text{if } y_{i\to j}^p = 1 \text{ and } x_i \neq x_j\\
0 & \text{otherwise}
\end{cases}
\end{equation}
where $\alpha_p$ is the positive penalty. Since $y_{i\to j}^p=1$ means an open positive gate between tracklet $i$ and $j$, it generates positive information to nodes $i$ and $j$ if $x_i$ and $x_j$ take different labels.

Similarly, the negative triplet potential is defined as
\begin{equation}
\small
\psi_t^n(x_i, x_j, y_{i\to j}^n) =\begin{cases}
\alpha_n & \text{if } y_{i\to j}^n = 1 \text{ and } x_i = x_j\\
0 & \text{otherwise}
\end{cases}
\end{equation}
where $\alpha_n$ is the negative penalty. Since $y_{i\to j}^n=1$ means an open negative gate between tracklet $i$ and $j$, it generate negative information to nodes $i$ and $j$ if $x_i$ and $x_j$ have the same label.

\subsection{Model Inference}\label{sec:label_inference}
% Directly looking for the MAP labeling that maximizes $P(\mathbf{x}, \mathbf{y})$ is a combinatorial optimization problem which is intractable. Instead, similar to \cite{crf2011}, we use mean field approximation to approximate the distribution $P(\mathbf{X}, \mathbf{Y})$ by the product of independent marginals
Directly looking for the label assignment that minimizes $E(\mathbf{x}, \mathbf{y})$ is a combinatorial optimization problem which is intractable. Instead, similar to \cite{crf2011}, we use the mean field method to approximate the distribution $P(\mathbf{X}, \mathbf{Y})\propto\exp(-E(\mathbf{X}, \mathbf{Y}))$ by the product of independent marginals
\begin{equation}
\small
\vspace{-5pt}
Q(\mathbf{X}, \mathbf{Y})=\prod_iQ_i(X_i)\prod_{j\in\mathcal{N}(i)}Q_{i\to j}^p(Y_{i\to j}^p)Q_{i\to j}^n(Y_{i\to j}^n)
\vspace{-5pt}
\end{equation}
Here $Q_i(X_i)$ is the identity distribution of node $i$, $Q_{i\to j}^p(Y_{i\to j}^p)$ and $Q_{i\to j}^n(Y_{i\to j}^n)$ are the status distributions of positive and negative gates on edge $i\to j$ respectively.

Let $\mathbf{q}_i^{(t)}=\begin{bmatrix}Q_i(1)^{(t)}&\cdots&Q_i(C)^{(t)}\end{bmatrix}^T$ be the identity distribution vector of node $i$ at the $t$-th iteration. $\pi_{i\to j}^{p, (t)} = Q_{i\to j}^{p, (t)}(1)$ and $\pi_{i\to j}^{n, (t)} = Q_{i\to j}^{n, (t)}(1)$ be the probability of opened positive and negative gates on edge $i\to j$ respectively.
Minimizing the KL-divergence $D(Q||P)$ between $P(\mathbf{X}, \mathbf{Y})$ and $Q(\mathbf{X}, \mathbf{Y})$ yields the following message passing updates:

\textbf{1)} For \textit{sample nodes}, we have
\vspace{-5pt}
{\small
\begin{align}\label{eq:tracklet_iteration}
\mathbf{q}_i^{(0)}=\operatorname{softmax}(T_{gt}\mathbf{S}_{:,i}^{gt}&)\nonumber\\
\mathbf{q}_i^{(t)}=\operatorname{softmax}(T_{gt}\mathbf{S}_{:,i}^{gt}&+\alpha_p\sum_{j\in\mathcal{N}(i)}\pi_{i\to j}^{p, (t-1)}\mathbf{q}_j^{(t-1)}\nonumber\\
&-\alpha_n\sum_{j\in\mathcal{N}(i)}\pi_{i\to j}^{n, (t-1)}\mathbf{q}_j^{(t-1)})
\vspace{-5pt}
\end{align}}%
where $\mathbf{S}_{:,i}^{gt}$ is the $i$th column of $\mathbf{S}^{gt}$.
% \begin{align}\label{eq:tracklet_iteration}
% Q_i^{(0)}(x_i=l)=&\frac{1}{Z_i}\exp\{T_{gt}s_{li}^{gt}\}\nonumber\\
% Q_i^{(t)}(x_i=l)=&\frac{1}{Z_i}\exp\{T_{gt}s_{li}^{gt} \nonumber\\
% &+\alpha_p\sum_{j\in\mathcal{N}(i)}Q_{i\to j}^{p, (t-1)}(1)Q_j^{(t-1)}(l)\nonumber\\
% &-\alpha_n\sum_{j\in\mathcal{N}(i)}Q_{i\to j}^{n, (t-1)}(1)Q_j^{(t-1)}(l)\}
% \end{align}
% where $Z_i$ is the normalization factor and $Q^{(t)}(\cdot)$ is the approximated distribution at the $t$th iteration. 

\textbf{2)} For \textit{gate nodes}, 
% we have
% \begin{align}
% &Q_{i\to j}^{p, (t)}(y_{i\to j}^p=1)\nonumber\\
% =&\frac{1}{Z_{i\to j}^p}\exp\{T_{tt}s_{ij}^{tt}+\alpha_p\sum_{l}Q_i^{(t-1)}(l)Q_j^{(t-1)}(l)-\alpha_p\}
% \end{align}
we let the marginal distribution of positive gates %set the factor $Z_{i\to j}^p$ so that 
$\sum_{j\in\mathcal{N}(i)}\pi_{i\to j}^{p, (t)}=1$ for normalization purpose. Then we have
\vspace{-5pt}
{\small
\begin{align}\label{eq:relation_iteration}
\pi_{i\to j}^{p, (0)}=&\operatorname*{softmax}_{\mathcal{N}(i)}(T_{tt}s_{ij}^{tt})\nonumber\\
\pi_{i\to j}^{p, (t)}=&\operatorname*{softmax}_{\mathcal{N}(i)}(T_{tt}s_{ij}^{tt}+\alpha_p\mathbf{q}_i^{(t-1)}\cdot\mathbf{q}_j^{(t-1)})
\vspace{-5pt}
\end{align}}%
where $\operatorname*{softmax}_{\mathcal{N}(i)}(\cdot)$ is the softmax operation in the neighborhood $\mathcal{N}(i)$.
% We also have
% \begin{align}
% &Q_{i\to j}^{n, (t)}(y_{i\to j}^n=1)=\frac{1}{Z_{i\to j}^n}\exp\{-\psi_u^n(1)\nonumber\\
% &-\sum_{l, l'}\sum_{j\in\mathcal{N}(i)}\psi_t^n(l, l', 1)Q_i^{(t-1)}(l)Q_j^{(t-1)}(l')\}\nonumber\\
% &=\frac{1}{Z_{i\to j}^n}\exp\left\{-\psi_u^n(1)-\alpha_n\sum_{l}Q_i^{(t-1)}(l)Q_j^{(t-1)}(l)\right\}
% \end{align}
From \eqref{eq:negative_relation}, we also have
\begin{equation}
\small
\vspace{-5pt}
\pi_{i\to j}^{n, (t)}=L_{ij}^{tt}
\end{equation}
for $t=0,\dots,K$. Thus, the marginal probability of a negative gate is fixed during inference.

% Before the recursive updating starts, these marginal probabilities are initialized as
% \begin{align}
% Q_i^{(0)}(x_i)&=\frac{1}{Z_i}\exp\{T_{gt}s_{x_i,i}^{gt}\}\nonumber\\
% Q_{i\to j}^{p, (0)}(y_{i\to j}^p=1)&=\frac{1}{Z_{i\to j}^p}\exp\{T_{tt}s_{ij}^{tt}\}
% \end{align}

%An example of message passing and node updating is shown in Figure~\ref{fig:message}.
From these recursive updating equations we can see that:

\textbf{1) }When updating \textit{sample node} $i$, identity information from $\mathbf{q}_j$ in $\mathcal{N}(i)$ is propagated through positive gate $\pi_{i\to j}^{p}$ and negative gate $\pi_{i\to j}^{n}$ and collected as positive ($\alpha_p$) and negative ($-\alpha_n$) message, respectively. These messages together with the prior identity information $\mathbf{S}_{:,i}^{gt}$ are combined to update $\mathbf{q}_i$, the identity distribution of node $i$, in the next iteration.

\textbf{2) }When updating \textit{gate node} $i\to j$, the identity similarity between $\mathbf{q}_i$ and its neighbor $\mathbf{q}_j$ in $\mathcal{N}(i)$ is measured by pairwise inner product. By combining this similarity with the initial contextual connection score $s_{ij}^{tt}$, the probability of gate openness $\pi_{i\to j}^{p}$ for the positive gate is updated. If $\mathbf{q}_i\cdot\mathbf{q}_j$ is small, $\pi_{i\to j}^{p}$ will gradually vanish in iterations, which avoids misleading connections propagating erroneous information. Negative gates based on cannot-links are fixed during inference.

% \textbf{1) }When updating \textit{sample node} $i$, identity distributions $\mathbf{q}_j$ in the neighborhood $\mathcal{N}(i)$ are propagated through positive and negative gates, weighted by the corresponding open-gate probabilities $\pi_{i\to j}^{p}$, $\pi_{i\to j}^{n}$ and collected as positive ($\alpha_p$) and negative ($-\alpha_n$) messages, respectively. These messages together with the prior identity information $\mathbf{S}_{:,i}^{gt}$ are combined and update $\mathbf{q}_i$, the identity distribution of node $i$, in the next iteration.

% \textbf{2) }When updating \textit{gate node} $i\to j$, the inner product between identity distributions $\mathbf{q}_i$ and $\mathbf{q}_j$ from the tracklet pair $(i, j)$ is used to measure the similarity of the identity distributions. By combining this similarity with the prior contextual connection score $s_{ij}^{tt}$, and compare with other edges connected to $i$, the open-gate probability $\pi_{i\to j}^{p}$ of the positive gate is updated. If $\mathbf{q}_i\cdot\mathbf{q}_j$ is small, $\pi_{i\to j}^{p}$ will probably drop after update, which avoids misleading connections propagating erroneous information. Negative gates based on cannot-links are fixed during inference.

We conduct these bidirectional updates jointly so that the samples nodes receive useful information from their neighbors through reliable connections to gradually refine their identity distributions, and the misleading connections in the graph are gradually corrected by these refined identity distributions in return. Please refer to the Supplementary Material for derivation details and illustrations of node update.
% It is easy to see that by minimizing the energy $E(\mathbf{x})$, the approach tends to assign each tracklet to the gallery subject with large facial similarity $s_{x_i,i}^F$. But different from the trivial solution, it also considers the correlation between tracklets since the tracklets that are similar in body appearance tend to be assigned to the same label.

After obtaining the approximation $Q(\mathbf{X}, \mathbf{Y})$ that minimizes $D(Q||P)$ in $K$ iterations,
% the inference can be easily applied by
% \begin{align}
% \mathbf{x}^*&=\operatorname*{argmax}_{\mathbf{x}\in\mathcal{L}^N}P(\mathbf{X})\approx\operatorname*{argmax}_{\mathbf{x}\in\mathcal{L}^N}Q(\mathbf{X})\nonumber\\
% &=\operatorname*{argmax}_{\mathbf{x}\in\mathcal{L}^N}\prod_i Q_i(x_i)\nonumber\\
% &=\left\{\operatorname*{argmax}_{x\in\mathcal{L}}Q_i(x)\right\}_{i=1,\dots,N}
% \end{align}
we use the identity distribution $\mathbf{q}_i^{(K)}$ as the output similarity scores $\tilde{\mathbf{S}}_{:, i}^{gt}$ from tracklet $i$ to gallery subjects.

\subsection{UGG: Training and Testing Settings}\label{sec:training}
% Depending on the availability of video face training data, the UGG model can be used in the following three settings:
\textbf{Testing with UGG: }
For testing, the UGG module can be directly applied at inference time, where we compute input matrices $\mathbf{S}^{gt}$, $\mathbf{S}^{tt}$ and $\mathbf{L}^{tt}$ from the video, setting the hyperparameters in the UGG module. Then the module produces the output similarity $\tilde{\mathbf{S}}^{gt}$ by recursive forward calculations.

\textbf{Training with UGG: }
Similar to RNN, the proposed UGG module can be considered as a differentiable recurrent module and be inserted into any neural networks for end-to-end training. If video face training data is available, we can utilize them for training to further improve the performance.

Given tracklets $\{T_i\}$ from a training video and galleries $\{G_l\}$, we use two DCNN networks $F_{gt}$ and $F_{tt}$ with parameters $\boldsymbol{\theta}_{gt}$ and $\boldsymbol{\theta}_{tt}$ pretrained on still images to generate $\mathbf{S}^{gt}$ and $\mathbf{S}^{tt}$ respectively as
\vspace{-5pt}
\begin{equation}\label{eq:input}
\small
s_{li}^{gt}=F_{gt}(G_l, T_i;\boldsymbol{\theta}_{gt}),\quad s_{ij}^{tt}=F_{tt}(T_i, T_j;\boldsymbol{\theta}_{tt})
\end{equation}
and feed into the UGG module.

After the module generates output similarity $\tilde{\mathbf{S}}^{gt}=\begin{bmatrix}\tilde{\mathbf{s}}_1,\dots,\tilde{\mathbf{s}}_N\end{bmatrix}$ after $K$ iterations, we compute the loss of this video as
\vspace{-5pt}
\begin{equation}\label{eq:loss}
\small
L = \frac{1}{N}\sum_{i\in \mathcal{S}}L_C(\tilde{\mathbf{s}}_i, z_i^c) + \lambda\frac{1}{N^2} \sum_{i, j\in\mathcal{S}}L_P(s_{ij}^{tt}, z_{ij}^b)
\vspace{-5pt}
\end{equation}
Here, $L_C$ is a cross-entropy loss on $\tilde{\mathbf{s}}_i$ with ground truth classification label $z_i^c$. % if tracklet $i$ and gallery $l$ have the same identity.
$L_P$ is a pairwise binary cross-entropy loss on $s_{ij}^{tt}$ with ground truth binary label $z_{ij}^b$. % = 1$ if tracklet $i$ and $j$ belongs to the same identity, and $z_{ij}^b=0$ otherwise.
%We use $L_C$ to encourage the output of the system to be discriminative so that given any tracklet, its similarity to the gallery with the same identity should be large compared to other galleries. $L_P$ is used to encourage the connection to be strong for positive pairs, and weak for negative pairs. %the tracklet pairs with the same identity to be more similar and those with different identities to be less similar.
$\lambda$ is the weight factor. $\mathcal{S}$ is the set of labeled tracklets.

Back-propagation through the whole networks on the overall loss $L$ is used to learn the DCNN parameters $\boldsymbol{\theta}_{gt}$, $\boldsymbol{\theta}_{tt}$ in $F_{gt}$ and $F_{tt}$, together with the temperature parameters $T_{gt}$, $T_{tt}$ in the UGG module. $T_{gt}$, $T_{tt}$ are learned in order to find a good balance between the unary scores and the messages from the neighbors during updates.

Depending on the different choices of $\mathcal{S}$, the training can be categorized into three settings:

\textbf{1. Supervised Setting:}
$\mathcal{S}=\mathcal{V}$, where every training sample in the graph is labeled. In this setting, we can directly utilize all the tracklets in the graph for training. 

\textbf{2. Semi-Supervised Setting:}
$\varnothing \subset \mathcal{S} \subset \mathcal{V}$, where training samples in the graph are only partially labeled. In this setting, the output of the module still depends on all the tracklets in the graph through information propagation. Thus, via back-propagation, the supervision information is propagated from labeled tracklets to unlabeled tracklets through the connections in the UGG module and enable them to benefit the training.

\textbf{3. Unsupervised Setting:}
$\mathcal{S}=\varnothing$, where no labeled training data is available. In this setting, we skip the training part since no supervision is provided.

%Related experiment results are discussed in Section~\ref{sec:semi}.
\section{Experiments}\label{sec:exp}
In this section, we report experiment results of the proposed method in two challenging video-based person search and face recognition datasets: the Cast Search in Movies (CSM) dataset \cite{personsearch} and the IARPA Janus Surveillance Video Benchmark (IJB-S) dataset \cite{ijbs}. %The state-of-the-art performance demonstrates the effectiveness of the proposed method.
\subsection{Datasets}
% We describe the details of both datasets as follows.
\textbf{CSM:} The CSM dataset is a large-scale person search dataset comprising a query set containing cast portraits in still images and a gallery set containing tracklets collected from movies. The evaluation metrics of the dataset include \emph{mean Average Precision} (mAP) and recall of the tracklet identification (R@k). Two protocols are used in the CSM dataset. One is IN which only search among tracklets in a single movie once a time. Another is ACROSS which search among tracklets in all the movies in the testing set. Please refer \cite{personsearch} for more details.%Notice that the query and gallery sets in a person search problem corresponds to the gallery and probe sets in a face recognition problem, respectively. In the following sections, we will follow the convention of face recognition.

\begin{table*}[ht]
\centering\resizebox{0.65\textwidth}{!}
{\begin{tabular}{|c|c|c|c|c|c|c|c|c|}
\hline
\multirow{2}{*}{Methods} & \multicolumn{4}{c|}{IN} & \multicolumn{4}{c|}{ACROSS}\\ 
\cline{2-9}
& mAP & R@1 & R@3 & R@5 & mAP & R@1 & R@3 & R@5\\
\hline
FACE(avg) & 53.33\% & 76.19\% & 91.11\% & 96.34\% & 42.16\% & 53.15\% & 61.12\% & 64.33\%\\
% \hline
% LD(avg) & 8.19\% & 39.70\% & 70.11\% & 87.34\% & 0.37\% & 0.41\% & 1.60\% & 5.04\%\\
\hline
PPCC(avg)\cite{personsearch} & 62.37\% & 84.31\% & 94.89\% & 98.03\% & 59.58\% & 63.26\% & 74.89\% & 78.88\%\\
\hline
PPCC(max)\cite{personsearch} & 63.49\% & 83.44\% & 94.40\% & 97.92\% & 62.27\% & 62.54\% & 73.86\% & 77.44\%\\
\hline
\hline
UGG-U(avg) & 62.81\% & 85.21\% & 95.65\% & 98.30\% & 63.31\% & 66.73\% & 76.09\% & 79.32\%\\
\hline
UGG-U(max) & 63.74\% & 84.93\% & 95.36\% & \textbf{98.37\%} & 63.42\% & 65.72\% & 74.90\% & 77.88\%\\
\hline
\hline
% RG(favg) & 63.88\% & 84.74\% & 94.95\% & 98.03\% & 64.44\% & 66.65\% & 74.94\% & 77.86\%\\
UGG-U(favg) & 64.36\% & 84.96\% & 94.90\% & 97.98\% & 64.85\% & 67.33\% & 75.38\% & 78.21\%\\
\hline
% UGG-U(favg-t) & 64.55\% & 86.34\% & 95.44\% & 98.25\% & 66.72\% & 70.53\% & 77.43\% & 79.92\%\\
% \hline
UGG-ST(favg) & 65.12\% & 86.73\% & 95.70\% & 98.34\% & 67.00\% & 71.16\% & 77.82\% & 80.15\%\\
\hline
%UGG-T(favg) & \textbf{65.47\%} & \textbf{87.21\%} & \textbf{95.89\%} & \textbf{98.37\%} & \textbf{67.29\%} & \textbf{71.72\%} & \textbf{78.28\%} & \textbf{80.49\%}\\
UGG-T(favg) & \textbf{65.41\%} & \textbf{87.28\%} & \textbf{95.87\%} & 98.28\% & \textbf{67.60\%} & \textbf{71.51\%} & \textbf{78.33\%} & \textbf{80.56\%}\\
\hline
\end{tabular}}
\vspace{-5pt}
\caption{Results on CSM dataset. Notice that \textit{UGG-U(favg)} is the unsupervised, initial setting before training. \textit{UGG-ST(favg)} is the semi-supervised training setting with 25\% samples labeled. \textit{UGG-T(favg)} is the supervised training setting.}% In \textbf{UGG-U(favg-t)}, the embeddings are trained without the UGG module.}
\label{tab:CSM}
\vspace{-5pt}
\end{table*}

% \begin{table*}[ht]
% \centering\resizebox{0.8\textwidth}{!}
% {\begin{tabular}{|c|c|c|c|c|c|c|c|c|}
% \hline
% \multirow{2}{*}{Methods} & \multicolumn{4}{c|}{IN} & \multicolumn{4}{c|}{ACROSS}\\ 
% \cline{2-9}
% & mAP & R@1 & R@3 & R@5 & mAP & R@1 & R@3 & R@5\\
% \hline
% FACE(avg) & 53.33\% & 76.19\% & 91.11\% & 96.34\% & 42.16\% & 53.15\% & 61.12\% & 64.33\%\\
% \hline
% LD(avg) & 8.19\% & 39.70\% & 70.11\% & 87.34\% & 0.37\% & 0.41\% & 1.60\% & 5.04\%\\
% \hline
% PPCC(avg) & 62.37\% & 84.31\% & 94.89\% & 98.03\% & 59.58\% & 63.26\% & 74.89\% & 78.88\%\\
% \hline
% PPCC(max) & 63.49\% & 83.44\% & 94.40\% & 97.92\% & 62.27\% & 62.54\% & 73.86\% & 77.44\%\\
% \hline
% \hline
% CRF(avg) & 62.43\% & 84.88\% & 95.31\% & 98.09\% & 62.31\% & 65.15\% & 75.68\% & 79.03\%\\
% \hline
% CRF(max) & 63.69\% & 84.52\% & 95.23\% & \textbf{98.26\%} & 60.03\% & 62.80\% & 74.42\% & 77.93\%\\
% \hline
% \hline
% CRF(favg) & 63.46\% & 84.26\% & 94.70\% & 97.92\% & 62.05\% & 64.44\% & 74.20\% & 77.59\%\\
% \hline
% CRF(train) & \textbf{64.05\%} & \textbf{86.64\%} & \textbf{95.60\%} & 98.24\% & \textbf{63.67\%} & \textbf{68.23\%} & \textbf{76.74\%} & \textbf{79.54\%}\\
% \hline
% \end{tabular}}
% \caption{Results on CSM dataset}
% \label{tab:CSM}
% \end{table*}

\begin{table*}[ht]
\centering\resizebox{\textwidth}{!}
{\begin{tabular}{|c|c|c|c|c|c|c|c|c|c|c|c|c|}
 \hline
  \multirow{2}{*}{Methods} & \multicolumn{6}{c|}{Top-K Average Accuracy \textit{with Filtering}} & \multicolumn{6}{c|}{EERR metric \textit{without Filtering}}\\ 
 \cline{2-13}
 & R@1 & R@2 & R@5 & R@10 & R@20 & R@50 & R@1 & R@2 & R@5 & R@10 & R@20 & R@50\\
 \hline
FACE(favg) & 64.86\% & 70.87\% & 77.09\% & 81.53\% & 86.11\% & 93.24\% & 29.62\% & 32.34\% & 35.60\% & 38.36\% & 41.53\% & 46.78\%\\
% \hline
% LD(favg) & 3.44\% & 5.77\% & 12.22\% & 20.83\% & 34.44\% & 63.19\% & 1.71\% & 2.89\% & 5.75\% & 9.46\% & 15.73\% & 30.31\%\\
\hline
PPCC(favg)\cite{personsearch} & 67.31\% & 73.21\% & 79.06\% & 83.12\% & 87.38\% & 93.68\% & 30.57\% & 33.28\% & 36.53\% & 39.10\% & 42.00\% & 47.00\%\\
\hline
FACE(sub)\cite{video_journal} & 69.82\% & 75.38\% & 80.54\% & 84.36\% & 87.91\% & 94.34\% & 32.43\% & 34.89\% & 37.74\% & 40.01\% & 42.77\% & 47.60\%\\
\hline
\hline
%RG(favg) & 73.08\% & 77.11\% & 81.36\% & 84.53\% & 88.10\% & 93.74\% & 32.25\% & 34.80\% & 37.39\% & 39.72\% & 42.40\% & 47.14\%\\
UGG-U(favg) & 74.20\% & 77.67\% & 81.43\% & 84.54\% & 87.96\% & 93.62\% & 32.70\% & 35.04\% & 37.54\% & 39.79\% & 42.43\% & 47.10\%\\
\hline
%RG(sub) & \textbf{76.31\%} & \textbf{80.25\%} & \textbf{83.79\%} & \textbf{86.44\%} & \textbf{89.21\%} & \textbf{94.74\%} & \textbf{33.87\%} & \textbf{36.57\%} & \textbf{39.00\%} & \textbf{40.91\%} & \textbf{43.26\%} & \textbf{47.85\%}\\
UGG-U(sub) & \textbf{77.59\%} & \textbf{80.46\%} & \textbf{83.70\%} & \textbf{86.20\%} & \textbf{89.23\%} & \textbf{94.55\%} & \textbf{34.79\%} & \textbf{36.88\%} & \textbf{39.11\%} & \textbf{40.90\%} & \textbf{43.37\%} & \textbf{47.86\%}\\
\hline
\end{tabular}}
\vspace{-5pt}
\caption{1:N Search results of IJB-S surveillance-to-single protocol. \textit{UGG-U(favg)} directly uses the cosine similarities between average-flattened features. \textit{UGG-U(sub)} uses the subspace-subspace similarity proposed in \cite{video_journal}.} %Top-K Average Accuracy is used for \textbf{with Filtering} and the EERR metric proposed in \cite{ijbs} is used for \textbf{without Filtering}.}
\label{tab:ijbs_sg2s}
\vspace{-15pt}
\end{table*}

\begin{table*}[ht]
\centering\resizebox{\textwidth}{!}
{\begin{tabular}{|c|c|c|c|c|c|c|c|c|c|c|c|c|}
 \hline
  \multirow{2}{*}{Methods} & \multicolumn{6}{c|}{Top-K Average Accuracy \textit{with Filtering}} & \multicolumn{6}{c|}{EERR metric \textit{without Filtering}}\\ 
 \cline{2-13}
 & R@1 & R@2 & R@5 & R@10 & R@20 & R@50 & R@1 & R@2 & R@5 & R@10 & R@20 & R@50\\
 \hline
FACE(favg) & 66.48\% & 71.98\% & 77.80\% & 82.25\% & 86.56\% & 93.41\% & 30.38\% & 32.91\% & 36.15\% & 38.77\% & 41.86\% & 46.79\%\\
% \hline
% LD(favg) & 3.41\% & 5.83\% & 12.75\% & 19.48\% & 33.30\% & 64.77\% & 1.83\% & 2.91\% & 5.83\% & 9.26\% & 15.20\% & 30.70\%\\
\hline
PPCC(favg)\cite{personsearch} & 68.96\% & 74.44\% & 79.84\% & 83.75\% & 87.68\% & 93.80\% & 31.37\% & 33.98\% & 37.04\% & 39.49\% & 42.35\% & 47.01\%\\
\hline
FACE(sub)\cite{video_journal} & 69.86\% & 75.07\% & 80.36\% & 84.32\% & 88.07\% & 94.33\% & 32.44\% & 34.93\% & 37.80\% & 40.14\% & 42.72\% & 47.58\%\\
\hline
\hline
%RG(favg) & 73.85\% & 77.94\% & 81.84\% & 84.99\% & 88.23\% & 93.91\% & 32.92\% & 35.36\% & 37.95\% & 40.05\% & 42.57\% & 47.16\%\\
UGG-U(favg) & 74.79\% & 78.35\% & 81.81\% & 84.85\% & 88.15\% & 93.80\% & 33.29\% & 35.48\% & 37.87\% & 40.02\% & 42.60\% & 47.14\%\\
\hline
%RG(sub) & \textbf{75.77\%} & \textbf{79.54\%} & \textbf{83.33\%} & \textbf{86.32\%} & \textbf{89.26\%} & \textbf{94.69\%} & \textbf{33.84\%} & \textbf{36.25\%} & \textbf{38.78\%} & \textbf{40.96\%} & \textbf{43.27\%} & \textbf{47.75\%}\\
UGG-U(sub) & \textbf{77.02\%} & \textbf{80.08\%} & \textbf{83.39\%} & \textbf{86.20\%} & \textbf{89.29\%} & \textbf{94.62\%} & \textbf{34.83\%} & \textbf{36.81\%} & \textbf{39.11\%} & \textbf{41.10\%} & \textbf{43.38\%} & \textbf{47.74\%}\\
\hline
\end{tabular}}
\vspace{-5pt}
\caption{1:N Search results of IJB-S surveillance-to-booking protocol. \textit{UGG-U(favg)} directly uses the cosine similarities between average-flattened features. \textit{UGG-U(sub)} uses the subspace-subspace similarity proposed in \cite{video_journal}.} %Top-K Average Accuracy is used for \textbf{with Filtering} and the EERR metric proposed in \cite{ijbs} is used for \textbf{without Filtering}.}
\label{tab:ijbs_b2s}
\vspace{-10pt}
\end{table*}

\textbf{IJB-S: } The IJB-S dataset is an unconstrained video face recognition dataset. The dataset is very challenging due to its low quality surveillance videos. In this paper, we mainly focus on two protocols related to our topic, the surveillance-to-single protocol (S2SG) and the surveillance-to-booking protocol (S2B). Galleries consist of single still image in S2SG and multiple still images in S2B. Probes are remotely captured surveillance videos from which all the tracklets are required. We report the per tracklet average top-K identification accuracy and the End-to-End Retrieval Rate (EERR) metric proposed in \cite{ijbs} for performance evaluation. Please refer \cite{ijbs} for more details.

\subsection{Implementation Details}\label{sec:implement}
%We describe the implementation details as follows.
\textbf{CSM:} For the CSM dataset, we use facial and body features provided by \cite{personsearch}. Please refer to the Supplementary Material for pre-processing details. Using the validation set, we choose parameters $T_{gt}=10$, $T_{tt}=15$, $\alpha_p=5$, $K=2$, $\lambda=0.1$ and $\lambda_f=0.1$ for the IN protocol and $T_{gt}=20$, $T_{tt}=30$, $\alpha_p=15$, $K=2$, $\lambda=0.1$ and $\lambda_f=0.1$ for the ACROSS protocol, in the UGG module for testing.

We also train linear embeddings on the provided features together with parameters in the UGG module in supervised settings. The training details are provided in the Supplementary Material.

\textbf{IJB-S:} For the IJB-S dataset, please refer to the Supplementary Material for pre-processing details. We empirically use the hyperparameter configuration of $T_{gt}=15$, $T_{tt} = 15$, $\alpha_p=10$, $\alpha_n=2$, $K=4$, $\lambda=0.1$ and $\lambda_f=0.1$ in the UGG module for testing.

To compare with \cite{video_journal}, we use the same configurations for tracklets filtering and evaluation metrics for each configuration: 1) \textit{with Filtering: }We keep those tracklets with length greater than or equal to 25 and average detection score greater than or equal to 0.9. 2) \textit{without Filtering}.
\subsection{Baseline Methods}
We conduct experiments on the CSM and IJB-S dataset with two baseline methods:
\textit{FACE:} facial similarity is directly used without any refinement. \textit{PPCC:} The Progressive Propagation via Competitive Consensus method proposed in \cite{personsearch} is used for post-processing. For the CSM dataset, we use the numbers reported in \cite{personsearch}. For the IJB-S dataset, we implement the method with code provided by the author.

For fair comparisons, following \cite{personsearch}, two settings of input similarity are used: \textit{avg:} similarity is computed by the average of all frame-wise cosine similarities between a gallery and a tracklet, or two tracklets. \textit{max:} similarity is computed by the maximum of all frame-wise cosine similarities between a gallery and a tracklet, or two tracklets. On IJB-S, we also implement the subspace-based similarity following \cite{video_journal}, denoted as \textit{sub}.

Two recent works \cite{cfan} and \cite{rean} have also reported results on the IJB-S dataset. These works built video templates by matching their detections with ground truth bounding boxes provided by the dataset. Our method follows \cite{video_journal} and associates faces across the video frames to build templates(tracklets) without utilizing any ground truth information. Since these two template building procedures are very different, a direct comparison is not meaningful.

Results of these baselines on two datasets are shown in Tables~\ref{tab:CSM}, \ref{tab:ijbs_sg2s} and \ref{tab:ijbs_b2s} respectively. Average run time of \textit{PPCC} is also reported in Table~\ref{tab:run_time}, on a machine with 72 Intel Xeon E5-2697 CPUs, 512GB of memory and two NVIDIA K40 GPUs. We observe that \textit{PPCC} only achieves marginal improvements on the IJB-S dataset. Its speed is also slow during inference, especially when large graphs are constructed.

\subsection{Evaluation on the Proposed UGG method}
On the CSM dataset, depending on the usage of training data, we evaluate three settings of UGG including: \textit{UGG-U:} without training, the UGG module works in \textit{unsupervised setting} as post-processing module. \textit{UGG-T:} with fully-labeled training data, the UGG module and linear embeddings are trained in \textit{supervised setting}. \textit{UGG-ST:} with 25\% labeled and 75\% unlabeled training data by random selection in each movie, the UGG module and linear embeddings and are trained in \textit{semi-supervised setting}. On the IJB-S dataset, since the dataset only provide test data, we use the \textit{unsupervised setting} and only test \textit{UGG-U}.

The additional input similarity used for training is the cosine similarity between flattened features after average pooling and denoted as \textit{favg}. Corresponding results are shown in Tables~\ref{tab:CSM}, \ref{tab:ijbs_sg2s} and \ref{tab:ijbs_b2s} respectively, with average run time tested on the same machine reported in Table~\ref{tab:run_time}.

{\textbf{Observations on CSM:}}

\textbf{1. UGG vs FACE:} All the settings of UGG perform significantly better than the raw baseline \textit{FACE}. \textit{UGG-T(favg)} provides state-of-the-art results on almost all the evaluation metrics with large margins, which demonstrates the effectiveness of the proposed method utilizing contextual connections.

\textbf{2. UGG vs PPCC \cite{personsearch}:} Using the same input similarity without training, \textit{UGG-U} performs better than \textit{PPCC} with relatively large margin, especially in the ACROSS protocol. Since in the ACROSS protocol, queries are searched among tracklets from all movies, the connections based on body appearance are not reliable across movies as those in the IN protocol. Thus by updating the gates between tracklets during inference, UGG is able to achieve much better performance than PPCC which is based on a fixed graph.

\textbf{3. Supervised vs Unsupervised:} From \textit{UGG-U(favg)} to \textit{UGG-T(favg)}, we observe significant improvements brought by training. It demonstrates that with labeled data, the UGG module can be inserted into deep networks for end-to-end training and achieve further performance improvement.

\textbf{4. Semi-Supervised vs Unsupervised:} We observe considerable improvements from \textit{UGG-U(favg)} to \textit{UGG-ST(favg)}. It implies that by reliable information propagation in the graphs, the UGG module can be trained with only partially-labeled data, and still achieves results comparable to the supervised setting.

\begin{table}[ht]
\centering\resizebox{0.33\textwidth}{!}
{\begin{tabular}{|c|c|c|c|c|}
 \hline
  \multirow{2}{*}{Methods} & \multicolumn{2}{c|}{CSM} & \multicolumn{2}{c|}{IJB-S}\\ 
 \cline{2-5}
 & IN & ACROSS & S2SG & S2B\\
  \hline
  PPCC\cite{personsearch} & 2.23s & 458.56s & 571.31s & 580.16s\\
  \hline
  UGG-U & 2.60s & 41.85s & 104.88s & 111.35s\\
 \hline
\end{tabular}}
\vspace{-5pt}
\caption{Average run time on CSM and IJB-S datasets.}
\label{tab:run_time}
\vspace{-10pt}
\end{table}

\begin{figure}[ht]
\centering
\includegraphics[width=\linewidth]{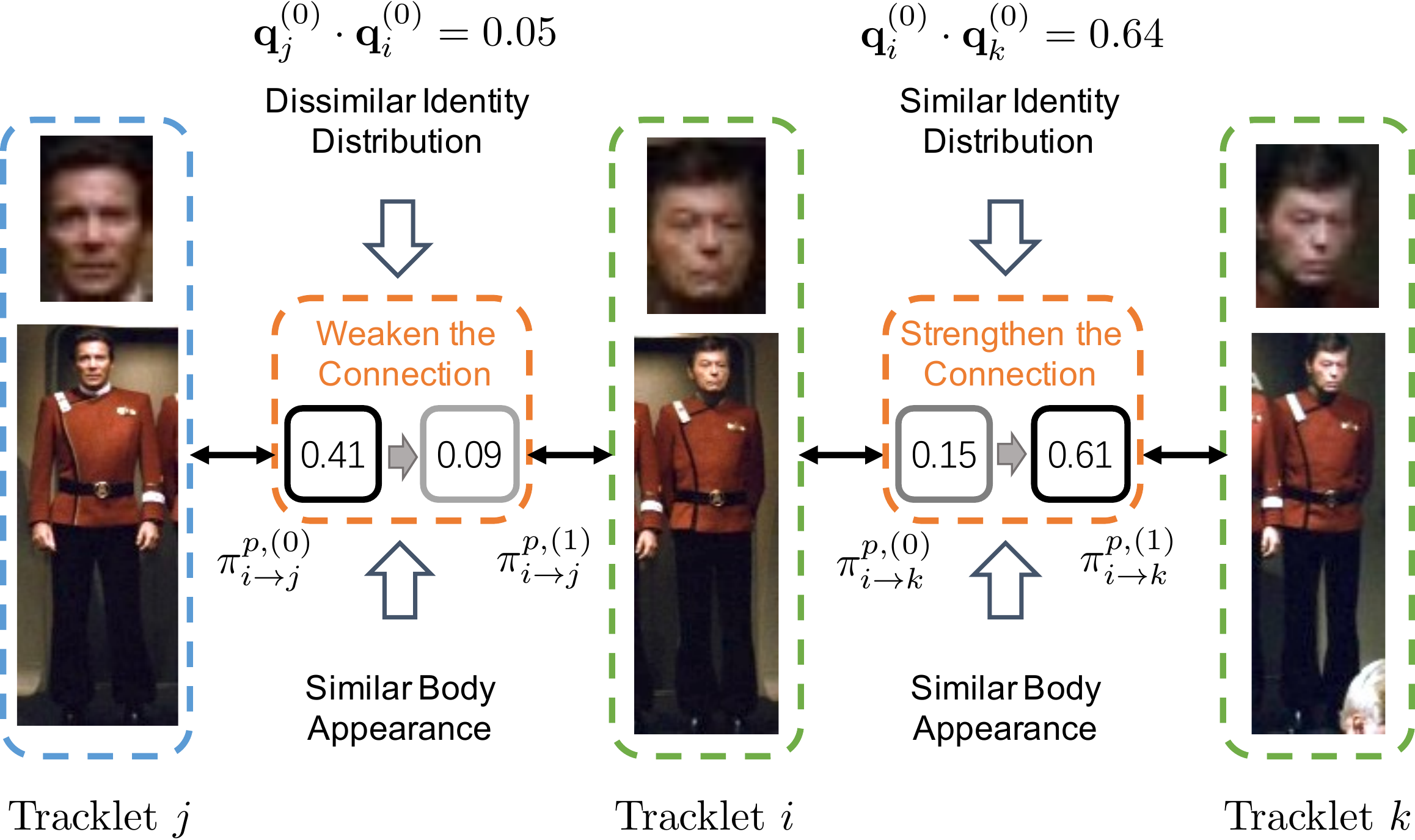}
\caption{A qualitative example from the CSM dataset. The positive connection between tracklets $i$ and $j$ is initially strong because of the similar body appearance. During the inference step of the proposed method, this connection is weakened because of the divergent identity distributions between the two tracklets. It avoids erroneous information propagation through the connection. In contrast, the connection between tracklets $i$ and $k$ is strengthened due to their similar identity distributions.}
\label{fig:connection}
\vspace{-15pt}
\end{figure}

{\textbf{Observations on IJB-S:}}

\textbf{1. UGG vs FACE and PPCC \cite{personsearch}:} \textit{UGG-U} performs better than \textit{FACE} and \textit{PPCC} on almost all evaluation metrics with relatively large margin, in both protocols, which again shows the effectiveness of the proposed method.

\textbf{2. UGG + Better Similarity Metric:} \textit{UGG-U(sub)} achieves state-of-the-art results by combining subspace-based similarity and UGG. It shows that the proposed method can further improve the performance over the improvement from the similarity metric.

\textbf{3. EERR Metric:} EERR metric \cite{ijbs} is relatively lower than identification accuracy, because it penalizes missed face detections,  which is out of the scope of this paper. 

{\textbf{Run time:}} From Table~\ref{tab:run_time}, we observe that UGG runs five times faster than PPCC on most of the protocols, which shows that UGG is more suitable for testing on large graphs during inference.

{\textbf{Qualitative Results:}} To illustrate the effectiveness of the proposed approach, a qualitative example is also shown in Figure~\ref{fig:connection}. Tracklets $i$ and $j$ belong to different identities and tracklets $i$ and $k$ belong to the same identity. The initialized positive gate probability {\small $\pi_{i\to j}^{p, (0)}=0.41$} is greater than {\small $\pi_{i\to k}^{p, (0)}=0.15$}. %Since the body similarity $s_{ij}^{tt}=0.74$ is greater than $s_{ik}^{tt}=0.67$, the initialized positive gate probability $\pi_{i\to j}^{p, (0)}=0.41$ is also greater than $\pi_{i\to k}^{p, (0)}=0.15$. 
If the gate is fixed, information will be erroneously propagated between $i$ and $j$. Using the proposed method, we can adaptively update the gate based on the identity information from $i$ and $j$. Since identity distribution similarity {\small $\mathbf{q}_j^{(0)}\cdot\mathbf{q}_i^{(0)}=0.05$} is very small, the two tracklets are unlikely to have the same identity. Hence the positive connection {\small $\pi_{i\to j}^{p, (1)}=0.09$} is weakened after the update. Similarly, since {\small $\mathbf{q}_i^{(0)}\cdot\mathbf{q}_k^{(0)}=0.64$} is large, the positive connection {\small $\pi_{i\to k}^{p, (1)}=0.61$} is strengthened correspondingly.% Then the information will propagate correctly in the next inference iteration.

\begin{table*}[ht]
\centering\resizebox{\textwidth}{!}
{\begin{tabular}{|c c c c|c|c|c|c|c|c|c|c|c|c|c|c|}
 \hline
  \multicolumn{4}{|c|}{\multirow{2}{*}{Configurations}} & \multicolumn{4}{c|}{CSM in avg} & \multicolumn{4}{c|}{CSM in max} & \multicolumn{4}{c|}{IJB-S in favg}\\ 
   \cline{5-16}
   \multicolumn{4}{|c|}{} &\multicolumn{2}{c|}{IN} & \multicolumn{2}{c|}{ACROSS} & \multicolumn{2}{c|}{IN} & \multicolumn{2}{c|}{ACROSS} & \multicolumn{2}{c|}{S2SG} & \multicolumn{2}{c|}{S2B}\\
 \hline
 PG & PGcl & NG & aG & mAP & R@1 & mAP & R@1 & mAP & R@1 & mAP & R@1 & A@1 & E@1 & A@1 & E@1\\
 \hline
 & & & & 58.72\% & 76.19\% & 55.67\% & 53.15\% & 61.29\% & 76.64\% & 58.20\% & 54.60\% & 64.86\% & 29.62\% & 66.48\% & 30.38\%\\
 \cline{5-16}
 \checkmark& & & & 61.14\% & 84.95\% & 62.00\% & 66.02\% & 61.60\% & 84.79\% & 62.05\% & 64.63\% & 71.21\% & 30.66\% & 72.05\% & 31.37\%\\
 \cline{5-16}
 & \checkmark& & & - & - & - & - & - & - & - & - & 71.26\% & 30.73\% & 72.16\% & 31.54\%\\
\cline{5-16}
& \checkmark& \checkmark& & - & - & - & - & - & - & - & - & 73.24\% & 32.35\% & 73.78\% & 32.88\%\\
\cline{5-16}
\checkmark& & & \checkmark& \textbf{62.81\%} & \textbf{85.21\%} & \textbf{63.30\%} & \textbf{66.73\%} & \textbf{63.74\%} & \textbf{84.93\%} & \textbf{63.42\%} & \textbf{65.72\%} & 72.32\% & 30.92\% & 73.15\% & 31.64\%\\
\cline{5-16}
 & \checkmark& & \checkmark& - & - & - & - & - & - & - & - & 72.46\% & 31.02\% & 73.28\% & 31.73\%\\
\cline{5-16}
& \checkmark& \checkmark& \checkmark & - & - & - & - & - & - & - & - & \textbf{74.20\%} & \textbf{32.70\%} & \textbf{74.79\%} & \textbf{33.29\%}\\
\hline
\end{tabular}}
\vspace{-5pt}
\caption{Ablation study. In configurations, \textit{PG} stands for adding positive gates for positive information. \textit{PGcl} stands for adding positive gates with extra control from cannot-links. \textit{NG} stands for adding negative gates for negative information. \textit{aG} stands for adaptively updating positive gates. \textit{A@1} stands for Average Accuracy \textit{with filtering} at R@1. \textit{E@1} stands for EERR \textit{without filtering} at R@1.}
\label{tab:ijbs_ablation}
\vspace{-8pt}
\end{table*}

\begin{table*}[ht]
\centering\resizebox{0.75\textwidth}{!}
{\begin{tabular}{|c c c|c|c|c|c|c|c|c|c|}
\hline
  \multicolumn{3}{|c|}{Configurations} & \multicolumn{4}{c|}{IN} & \multicolumn{4}{c|}{ACROSS} \\ 
   \hline
   PGTrain & aGTrain & UGGTest & mAP & R@1 & R@3 & R@5 & mAP & R@1 & R@3 & R@5\\
\hline
& & & 61.13\% & 77.86\% & 91.79\% & 96.65\% & 58.34\% & 56.56\% & 63.83\% & 66.34\%\\
\cline{4-11}
\checkmark& & & 61.39\% & 77.99\% & 91.77\% & 96.61\% & 58.94\% & 57.31\% & 64.26\% & 66.88\%\\
\cline{4-11}
\checkmark& \checkmark& & 61.40\% & 78.12\% & 91.85\% & 96.67\% & 58.70\% & 57.64\% & 64.49\% & 67.22\%\\
\cline{4-11}
&&\checkmark& 64.14\% & 85.90\% & 95.42\% & 98.10\% & 65.82\% & 69.45\% & 76.83\% & 79.34\%\\
\cline{4-11}
\checkmark& &\checkmark& 64.58\% & 86.36\% & 95.53\% & \textbf{98.27\%} & 66.90\% & 70.74\% & 77.83\% & 80.02\%\\
\cline{4-11}
\checkmark&\checkmark&\checkmark& \textbf{64.60\%} & \textbf{86.68\%} & \textbf{95.56\%} & 98.24\% & \textbf{67.09\%} & \textbf{71.31\%} & \textbf{77.93\%} & \textbf{80.39\%}\\
\hline
\end{tabular}}
\vspace{-5pt}
\caption{Additional study on semi-supervised training on CSM dataset. \textit{PGTrain} stands for using fixed positive gates during training. \textit{aGTrain} stands for adaptively updating the gates during training. \textit{UGGTest} stands for using UGG model during testing. In all experiments, only 25\% of the training samples are labeled.}
\label{tab:CSM_semi}
\vspace{-15pt}
\end{table*}

\subsection{Ablation Studies}
We conduct ablation studies on CSM and IJB-S datasets to show the effectiveness of key features in the proposed model. The results are shown in Table~\ref{tab:ijbs_ablation}. We start from the baseline \textit{FACE} without any information propagation, then gradually add key features of the method: \textit{PG:} add fixed positive gates to propagate positive information. \textit{PGcl:} same as \textit{PG} except that positive information will not be propagated when cannot-link exists. \textit{NG:} add negative gates to propagate negative information. \textit{aG:} adaptively update positive gates in \textit{PG} or \textit{PGcl} using the proposed method.
Since detection information is not given in the CSM dataset, there is no co-occurrence cannot-links available and we do not use negative gates in this dataset. Thus, the proposed method \textit{UGG-U} corresponds to \textit{PG}+\textit{aG} on the CSM dataset and \textit{PGcl}+\textit{NG}+\textit{aG} on the IJB-S dataset.

From Table~\ref{tab:ijbs_ablation} we observe that: \textbf{1)} by introducing fixed positive gates, the performance improves compared to the baseline results, which indicates that positive information propagation controlled by body similarity improves the performance. \textbf{2)} by adding cannot-links to control the positive gates as well, marginal improvements are obtained. Thus, the performance improvement is limited if allow only positive information to propagate. \textbf{3)} by introducing additional negative gates using the same cannot-links, the performance improves significantly, which demonstrates the effectiveness of allowing negative information to propagate between tracklets. \textbf{4)} finally, by adaptively updating the positive gates, we achieve the best performance in all protocols of both datasets. The result implies the advantages of adaptively updated gates.

\subsection{Experiments on Different Training Settings}\label{sec:semi}
We also perform additional experiments on semi-supervised training on the CSM dataset with results shown in Table~\ref{tab:CSM_semi}. In the experiment, similar to the \textit{UGG-ST} setting, we first randomly pick 25\% tracklets in each graph as labeled samples, and the rest 75\% as unlabeled. We only train the linear embedding on face features with fixed UGG module on these training data.

Suppose after applying the embedding we want to learn, the similarities between galleries and labeled/unlabeled tracklets are $\mathbf{S}^{gt}=\begin{bmatrix}\mathbf{S}^{gt}_l, \mathbf{S}^{gt}_u\end{bmatrix}$. We use three different settings to train the embedding: \textbf{1)} directly train on the labeled similarities $\mathbf{S}^{gt}_l$ using cross-entropy loss, without invoking the UGG module. \textbf{2)} use the UGG module with positive gates to process $\mathbf{S}^{gt}$ and train on the output similarity $\tilde{\mathbf{S}}^{gt}_l$ corresponding to the labeled tracklets by cross-entropy loss, denoted as \textit{PGTrain}. \textbf{3)} adaptively update the positive gates used in \textit{PGTrain}, denoted as \textit{aGTrain}. Please refer to the Supplementary Material for training details.

Two settings are used to test the performance of the embedding: \textbf{1)} directly test on $\mathbf{S}^{gt}$ from the learned embedding, without using the UGG as post-processing. \textbf{2)} test on $\tilde{\mathbf{S}}^{gt}$ from the learned embedding and with the UGG post-processing, denoted as \textit{UGGTest}.

From the results in Table~\ref{tab:CSM_semi}, we observe that in the semi-supervised setting, the embedding trained with the UGG is more discriminative than the one trained without the module. It achieves better performance in both test settings. It shows that by propagating information between tracklets, the UGG also leverages the information from those unlabeled tracklets during training, which is important for semi-supervised learning. Also, the UGG with adaptive gates performs better than fixed gates, which demonstrates that adaptive gates is also helpful during training by propagating the information more precisely between tracklets.
\section{Conclusions and Future Work}\label{sec:conclusion}
% In this paper, we proposed a  graphical-model-based framework called the Uncertainty-Gated Graph, for video-based face recognition problem. The framework propagates positive and negative identity information between samples through the adaptive connections implied by both the context information and the identity correlation between samples. %The framework propagate identity information from high-quality samples to low-quality samples through the connections implied from the context information. It uses gate nodes and the associated random fields to model the uncertainty of connections between samples and enable the edge weights to be adaptively adjusted according to the neighboring samples. The gates also allow both positive and negative information to propagate simultaneously.
% The proposed framework can be either used as post-processing, or trained in supervised and semi-supervised ways. It achieves state-of-the-art results on the CSM and the IJB-S datasets. Ablation study further shows the effectiveness of the key features of the proposed method. An interesting future work will be using the gender information to build negative connections and adaptively update negative gates as well.

In this paper, we proposed a graphical model-based method %, the Uncertainty-Gated Graph,
for video-based face recognition. The method propagates positive and negative identity information between tracklets through adaptive connections, which are influenced by both contextual information and identity distributions between tracklets. %The framework propagate identity information from high-quality samples to low-quality samples through the connections implied from the context information. It uses gate nodes and the associated random fields to model the uncertainty of connections between samples and enable the edge weights to be adaptively adjusted according to the neighboring samples. The gates also allow both positive and negative information to propagate simultaneously.
The proposed method can be either used for post-processing, or trained in supervised and semi-supervised fashions. It achieves state-of-the-art results on CSM and IJB-S datasets. %Ablation study further implies the effectiveness of the key features of the proposed method.
An interesting future work will be using attribute information, such as gender, to construct negative connections and adaptively update negative gates.% as well.
{\small
\section*{Acknowledgment}

This research is based upon work supported by the Office of the Director of National Intelligence (ODNI), Intelligence Advanced Research Projects Activity (IARPA), via IARPA R\&D Contract No. 2019-022600002. The views and conclusions contained herein are those of the authors and should not be interpreted as necessarily representing the official policies or endorsements, either expressed or implied, of the ODNI, IARPA, or the U.S. Government. The U.S. Government is authorized to reproduce and distribute reprints for Governmental purposes notwithstanding any copyright annotation thereon.

}

{\small
\bibliographystyle{ieee_fullname}
\bibliography{refs}
}

\end{document}